\pdfoutput=1

\documentclass[11pt]{article}

\usepackage{naacl2021}

\usepackage{times}
\usepackage{latexsym}

\usepackage[T1]{fontenc}

\usepackage[utf8]{inputenc}

\usepackage{microtype}

\usepackage{booktabs} 

\usepackage{blindtext}
\usepackage{multirow}
\usepackage{graphicx}

\usepackage{amsmath}
\usepackage{amssymb}
\usepackage{float}

\newcommand\magic[1]{\textbf{\textcolor{blue}{#1}}}

\title{Can Latent Alignments Improve Autoregressive Machine Translation?}

\author{Adi Haviv\thanks{\;\;~~Equal contribution.}$\ \hspace{0.05cm}$~~~~~~~~~Lior Vassertail$^{*}$~~~~~~~~~Omer Levy\\ \\
The Blavatnik School of Computer Science \\
Tel Aviv University \\
\small{\texttt{ \{adi.haviv,lior.vassertail,levyomer\}@cs.tau.ac.il }}
}

\begin{document}
\maketitle
\begin{abstract}
Latent alignment objectives such as CTC and AXE significantly improve non-autoregressive machine translation models.
Can they improve \textit{autoregressive} models as well?
We explore the possibility of training autoregressive machine translation models with latent alignment objectives,
and observe that, in practice, this approach results in degenerate models.
We provide a theoretical explanation for these empirical results, and prove that latent alignment objectives are incompatible with teacher forcing.
\end{abstract}


\section{Introduction}

Latent alignment objectives, such as CTC \cite{Graves2006ConnectionistTC} and AXE \cite{Ghazvininejad2020AlignedCE}, have been recently proposed for training non-autoregressive models for machine translation \cite{Libovick2018EndtoEndNN, saharia-etal-2020-non}.
These objectives use a dynamic program to comb the space of monotonic alignments between the ``gold'' target sequence and the token probabilities the model predicts, thus reducing the loss from positional misalignments and focusing on the original prediction error instead. 
For example, consider the target sequence \textit{``there is a tiny difference between pink and magenta''}; if the model's distribution favors the paraphrase \textit{``there is a very small difference between pink and magenta''}, substituting one token (\textit{``tiny''}) with two (\textit{``very small''}) will cause a misalignment, and result in a disproportionately large cross entropy loss. A latent alignment loss would match the predictions of both \textit{``very''} and \textit{``small''} with the target \textit{``tiny''}, while aligning the rest of the sentence properly and computing a much lower loss that focuses on this particular discrepancy.
Could latent alignments also benefit \textit{autoregressive} models?

We apply CTC and AXE to standard autoregressive machine translation models.
We observe that, when trained with teacher forcing \cite{Williams1989ALA}, CTC reduces to the vanilla cross entropy loss because CTC assumes that the prediction sequence is longer than the target and has only one valid alignment when they are equal.
We further examine AXE, which does not share this assumption, and find that it yields a degenerate model that almost perfectly fits the training set but completely fails at inference time.

Our analysis reveals that latent alignments and teacher forcing are fundamentally incompatible.
We observe that there exists a valid alignment in which the prediction $p_{i}$ is aligned with the target $y_{i-1}$ for almost every token.
Simultaneously, teacher forcing feeds the model with $y_{i-1}$ when computing the prediction $p_{i}$, encouraging the model to simply predict its input under this alignment.
While AXE allows this alignment for equal-length prediction and target sequences, the phenomenon also occurs (theoretically) in CTC if the predictions are longer, and in fact, occurs in any latent alignment objective that can align a prediction $p_j$ with a target $y_i$ where $i<j$.

\section{Background: Latent Alignments}
\label{sec:background}

A latent alignment objective measures the compatibility between the target sequence $Y$ and the sequence of predicted token probabilities $P$ by considering a subspace of possible mappings between $Y$ and $P$.
Latent alignments are typically used in non-autoregressive models for automatic speech recognition, and optical character recognition \cite{Graves2006ConnectionistTC}, and have recently been introduced to the task of machine translation \cite{Libovick2018EndtoEndNN, Ghazvininejad2020AlignedCE, saharia-etal-2020-non}.
We describe two such objectives, beginning with an overview of the common notation and framework.

\begin{table*}[t]
\small
\centering
\begin{tabular}{@{}lp{0.45\textwidth}lcc@{}}
\toprule
\textbf{Operator} & \textbf{Description} & \textbf{Formula} & \textbf{CTC} & \textbf{AXE} \\
\midrule
\textit{Align} & Predict the target token $Y_{i}$ with the distribution $P_{j}$. This is the default alignment, advancing along $A$'s diagonal. & $A_{i,j} = A_{i-1,j-1} \cdot P_j (Y_i) $ & \checkmark & \checkmark \\
\textit{Clone Target} & Assuming the target token $Y_{i}$ was predicted with the previous distribution $P_{j-1}$, repredict $Y_{i}$ using $P_{j}$. & $A_{i,j} = A_{i,j-1} \cdot P_j (Y_i)$ & \checkmark & \\
\textit{Clone Prediction} & Assuming the previous target token $Y_{i-1}$ was predicted with the distribution $P_{j}$, reuse $P_{j}$ to predict the next target token $Y_{i}$. & $A_{i,j} = A_{i-1,j} \cdot P_j (Y_i)$ & & \checkmark \\
\textit{Delimiter} & Use the distribution $P_{j}$ to predict the blank token $\varepsilon$ instead of the target token $Y_{i}$. This operation is akin to inserting $\varepsilon$ into the target sequence at the $i$-th position. & $A_{i,j} = A_{i,j-1} \cdot P_j (\varepsilon)$ & \checkmark & \checkmark \\
\bottomrule
\end{tabular}
\caption{The set of possible operators in latent alignment dynamic programs. Both AXE and CTC use \textit{align} and \textit{delimiter}, but apply different \textit{clone} operators.}
\label{tab:latent-alignment-operators}
\end{table*}

\paragraph{Monotonic Alignments}
Let $Y = y_1,\ldots,y_n$ be the target sequence of $n$ tokens, and $P = p_1,\ldots,p_m$ be the model prediction, a sequence of $m$ token probability distributions. 
A monotonic alignment $\alpha$ is a function that maps every target position $i \in \{1, \ldots , n\}$ to a set of one or more \textit{consecutive} prediction positions $\alpha(i) \subseteq \{1, \ldots , m\}$, such that  $i \le j \Leftrightarrow \max{\alpha(i)} \leq \min{\alpha(j)}$.

\paragraph{Objective}
Given an alignment $\alpha$, the objective is defined as follows:
\begin{align}
L_\alpha (Y, P) = \prod_{i=1}^{n}\prod_{j\in\alpha(i)}p_{j}(y_{i})
\label{eq:alignment-model}
\end{align}
Since $\alpha$ is not provided a priori, it is necessary to aggregate over all the possible alignments (hence \textit{latent} alignments), by either summation (Equation~\ref{eq:summation}) or maximization (Equation~\ref{eq:maximization}):
\begin{align}
L^{\sum} (Y, P) &= \sum_\alpha L_\alpha (Y, P) \label{eq:summation} \\
L^{\max} (Y, P) &= \max_\alpha L_\alpha (Y, P) \label{eq:maximization}
\end{align}
In practice, the negative log loss is minimized during training:
\begin{align}
\ell (Y, P) = -\log L (Y, P)
\end{align} 

\paragraph{Dynamic Programming}
Aggregation can be done efficiently with dynamic programming, using derivations of the forward-backward algorithm (for summation, as in CTC) or the Viterbi algorithm (for maximization, as in AXE).
These algorithms create an aggregation matrix $A \in R^{n \times m}$, where each cell represents the desired aggregation score $f$ (sum or max) over prefixes of the target and prediction probability sequences:
$A_{i,j} = L^{f}(Y_{\leq i}, P_{\leq j})$.
The dynamic program combs through the space of alignments by implicitly constructing every possibility using the set of local operators defined in Table~\ref{tab:latent-alignment-operators}.
The subspace of alignment functions that the program explores is determined by the subspace of operators it employs.

\paragraph{Connectionist Temporal Classification (CTC)}
The CTC objective \cite{Graves2006ConnectionistTC} was originally introduced for speech and handwriting recognition, where the prediction sequence $P$ is typically much longer than the target sequence $Y$ ($m \gg n$).
While computing the summation objective (Equation~\ref{eq:summation}), CTC uses only the \textit{align}, \textit{clone target}, and \textit{delimiter} operators.
This means that CTC restricts $\alpha$ to the space of alignments where every item in $P$ is aligned with at most one item in $Y$, i.e. $\alpha(i) \cap \alpha(j) = \emptyset$ for $i \neq j$.

CTC was used in non-autoregressive machine translation by \citet{Libovick2018EndtoEndNN} and more recently by \citet{saharia-etal-2020-non}.
In both cases, the prediction sequence was artificially inflated to be double (or more) the length of the source-language input sequence in order to simulate the $m \gg n$ condition of speech recognition.

\paragraph{Aligned Cross Entropy (AXE)}
The AXE objective \cite{Ghazvininejad2020AlignedCE} is specifically designed for non-autoregressive machine translation. AXE finds the monotonic alignment that minimizes the cross entropy loss (i.e., maximizes the likelihood, Equation~\ref{eq:maximization}) in order to focus the penalty on the root errors instead of positional shifts that result from them.
AXE uses only the \textit{align}, \textit{clone prediction}, and \textit{delimiter} operators.
This combination of operators allows AXE to align prediction and target sequences of any lengths because \textit{clone prediction} inflates the prediction sequence while \textit{delimiter} adds new target tokens.
However, since AXE cannot clone target tokens, every target position $i$ is always aligned to a \textit{single} prediction position, i.e. $|\alpha (i)| = 1$.
Figure~\ref{fig:alignment-example} illustrates how AXE aligns the model's predictions with the target sequence.

\begin{figure}[t]
\centering
\small
\begin{tabular}{@{}lccccc@{}} 
\toprule
\textbf{Target $Y$} & it & is & rainy & today & EOS\\
\midrule
\multirow{4}{5em}{\textbf{Model Predictions $P$
(Top 4)}} & \magic{it} & \magic{is} & so & \magic{rainy} & \magic{today} \\ 
  & however & the & rain & good & tonight \\ 
  & the & looks & very & and & \magic{EOS} \\ 
  & but & this & \magic{$\varepsilon$} & very & good \\ 
\bottomrule 
\end{tabular}
\caption{
An illustration of how AXE aligns the model's predictions $P$ with the target sequence $Y$: \textit{``it is rainy today''}.
The model favors a slightly different sequence (\textit{``it is so rainy today''}), which would suffer from a high penalty with the regular cross entropy loss. Instead, AXE finds a more appropriate alignment $\alpha = (1, 2, 4, 5, 5)$ using the operator sequence \textit{align}, \textit{align}, \textit{delimiter}, \textit{align}, \textit{align}, \textit{clone prediction}.
}
\label{fig:alignment-example}
\end{figure}
\section{Combining CTC with Teacher Forcing Defaults to the Trivial Alignment}

In an \textit{autoregressive} setting, it is standard practice to use teacher forcing \cite{Williams1989ALA}; i.e., when predicting the $i$-th token, the model takes the prefix of the (gold) target sequence $Y_{<i}$ as input.
This dictates that the number of predictions is identical to the number of target tokens ($m = |P| = |Y| = n$).

However, CTC assumes that the prediction sequence $P$ is typically much longer than the target sequence $Y$ ($m \gg n$), and can only inflate $Y$ via \textit{clone target} and \textit{delimiter} (see Section~\ref{sec:background}).
This leaves only one valid alignment when $m=n$: the \textit{trivial} alignment $\alpha(i) = \{i\}$.
CTC will thus default to the same objective as the standard cross entropy loss.

Unlike CTC, the AXE objective aggregates over multiple alignments even when $m=n$, because it uses both the \textit{delimiter} operator (which inflates $Y$) as well as the \textit{clone prediction} operator (which inflates $P$).

\section{Applying AXE to Autoregressive NMT}

To apply AXE to autoregressive machine translation, we use a standard sequence-to-sequence transformer model \cite{Vaswani2017AttentionIA} trained with teacher forcing, replace the simple cross entropy loss function with AXE, and add the empty token $\varepsilon$ to the vocabulary.
We remove the $\varepsilon$ tokens after decoding.

\paragraph{Experiment Setup}
We use \texttt{fairseq} \cite{ott2019fairseq} to train a transformer encoder-decoder \cite{Vaswani2017AttentionIA} on the IWSLT'14 DE-EN dataset \cite{Cettolo2015ReportOT}. The dataset is preprocessed and tokenized into subwords with BPE \cite{Sennrich2016NeuralMT} using the scripts provided by \texttt{fairseq}.
We also use the implementation's default hyperparameters: 6 layers of encoder/decoder, 512 model dimensions, 1024 hidden dimensions, 4 attention heads.
We optimize with Adam \cite{Kingma2015AdamAM} for 50k steps with early stopping using 4096 tokens per batch.
We decode with beam search ($b=5$) and evaluate performance with BLEU \cite{papineni-etal-2002-bleu}.

\begin{figure}
\centering
\includegraphics[width=1.0\linewidth]{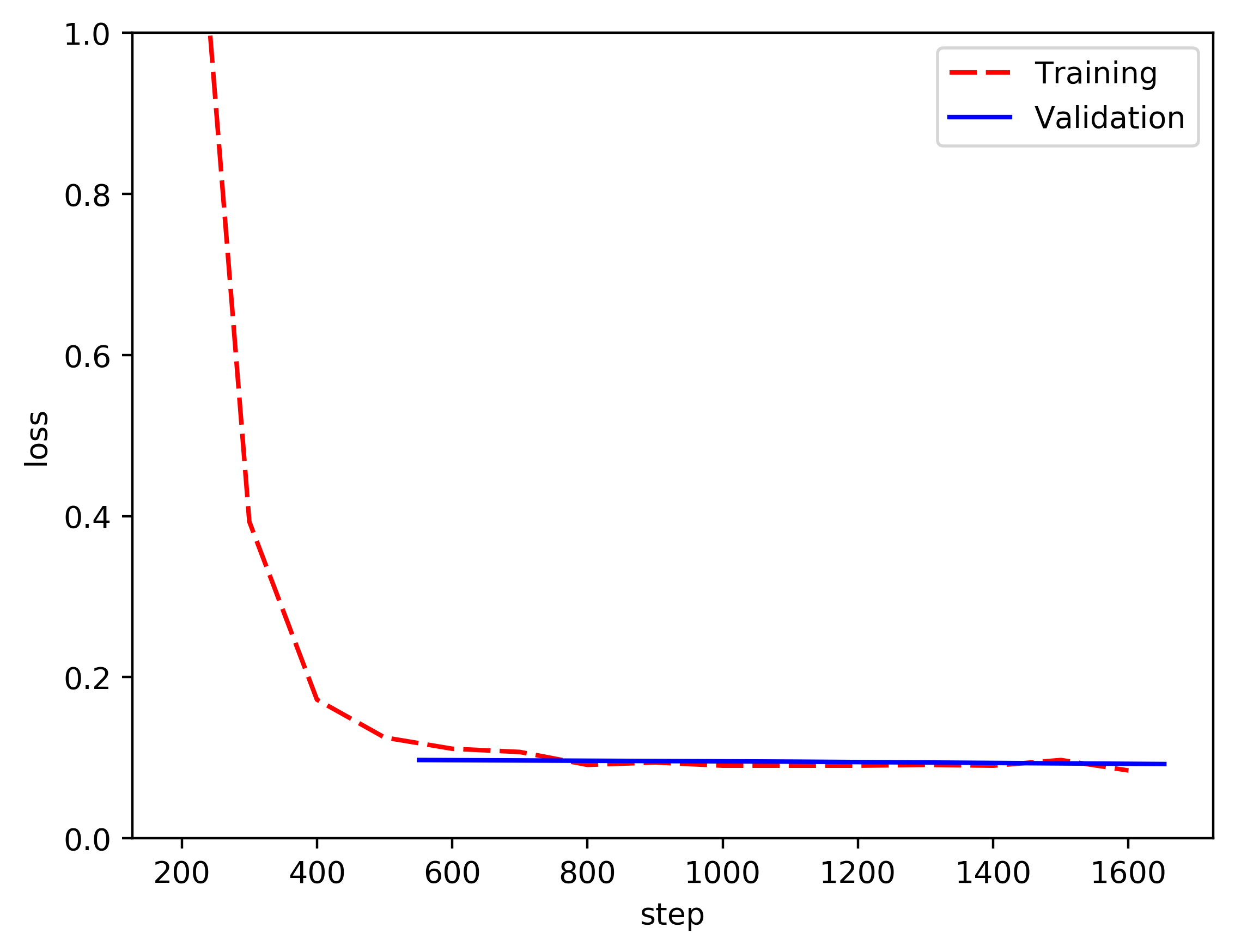}
\caption{Training and validation loss when using the AXE objective on IWSLT'14 DE-EN with an autoregressive model.}
\label{fig:axe_loss}
\end{figure}

\paragraph{Results}
We observe two seemingly contradictory behaviors.
On the one hand, the model approaches a near-zero training loss within a single epoch, and observes similar results when computing AXE loss on unseen examples in the validation set (Figure~\ref{fig:axe_loss}).
Meanwhile, at inference time, the model consistently produces the empty sequence (after removing all instances of $\varepsilon$), scoring 0 BLEU on the test set.
This indicates that the model has learned to ``game'' the AXE objective without actually learning anything useful about machine translation.
What shortcut did the model learn?

\section{Analysis}

\begin{figure*}[t]
\footnotesize
\begin{center}
\begin{tabular}{@{}lr@{~~}lr@{~~}lr@{~~}lr@{~~}lr@{~~}lr@{~~}l@{}}
\toprule
\textbf{Prediction $Y$} & \multicolumn{2}{c}{thank} & \multicolumn{2}{c}{you} & \multicolumn{2}{c}{for} & \multicolumn{2}{c}{listening} & \multicolumn{2}{c}{.} & \multicolumn{2}{c}{EOS} \\
\midrule
\textbf{Alignment $\alpha(i)$} & \multicolumn{2}{c}{2} & \multicolumn{2}{c}{3} & \multicolumn{2}{c}{4} & \multicolumn{2}{c}{5} & \multicolumn{2}{c}{6} & \multicolumn{2}{c}{6} \\
\midrule
\multirow{4}{5em}{\textbf{Model Predictions $P$
(Top 4)}} & 0.999 & \magic{$\varepsilon$} & 0.995 & \magic{thank} & 0.999 & \magic{you} & 0.999 & \magic{for} & 0.994 & \magic{listening} & 0.627 & \magic{.} \\
& 8e-8 & EOS & 5e-5 & 's & 2e-7 & pre@@ & 1e-5 & is & 3e-5 & ver@@ & 0.370 & \magic{EOS} \\ 
& 8e-8 & ... & 2e-5 & super@@ & 2e-7 & ke & 6e-6 & audience & 2e-5 & taking & 1e-4 & ... \\ 
& 7e-8 & use & 2e-5 & unfortunate & 2e-7 & cu@@ & 5e-6 & oil & 2e-5 & sever@@ & 1e-4 & ' \\  
\bottomrule
\end{tabular}
\caption{An example of the constant alignment that AXE chooses after training the model. Given the German source \textit{``danke fürs zuhören''}, the model tries to predict \textit{``thank you for listening''}. Because the model is trained with teacher forcing, it can simply learn to predict its input at each position, and assume that AXE will align the prediction with the previous token (which is identical to the input). For example, $p_2$ predicts \textit{``thank''} with very high probability because teacher forcing uses the previous target $y_1$ as the decoder's input in the second position. Notice how the final prediction $p_6$ is used twice to predict both \textit{``.''} and \textit{EOS}.}
\label{fig:sample-axe-path}
\end{center}
\end{figure*}

To understand \textit{how} the model learns to game the AXE objective, we analyze the optimal alignments chosen by the objective, and find that they allow the model to condition on the target token when trying to predict it.
We prove that this is the optimal solution when combining teacher forcing and AXE, and that it holds for any latent alignment objective that allows the model to align future target tokens with the current prediction.

\paragraph{AXE finds a constant alignment} 
We examine the alignments chosen by AXE's dynamic program for a sample of training examples, and observe that they all belong to a consistent pattern: \textit{delimiter}, \textit{align}, \textit{align}, ..., \textit{clone prediction}.
In other words, the chosen path skips the first prediction by emitting the blank token $\varepsilon$ and then aligns each prediction $p_i$ with the \textit{previous} target token $y_{i-1}$.
The alignment synchronizes the positions at the end of the sequence by cloning the last prediction to compensate for the offset produced by the initial \textit{delimiter} operator.

\paragraph{Each prediction conditions on its target}
The teacher forcing algorithm conditions the prediction $p_i$ on the ground truth of the previous tokens $y_1, \ldots, y_{i-1}$ to predict the target token $y_i$.
However, if the prediction $p_i$ is aligned with the target $y_{i-1}$, then it is effectively observing its target through the input, and only needs to learn the identity function.
Formally, we see that for every $1 < i < n$ the prediction is trivial:
\begin{align*}
p_i(y_{i-1}) &= Pr(y_{i-1}|X,Y_{<i}) \\
 &= Pr(y_{i-1}|y_{i-1}) = 1
\end{align*}
Figure~\ref{fig:sample-axe-path} demonstrates this phenomenon on an actual example using the model's predictions.

\paragraph{The cost of sharing the last prediction} 
It is now clear to see that the loss should indeed be close to zero.
Having said that, it is not infinitesimal; the last two tokens (typically \textit{``.''} and \textit{EOS}) need to be predicted from the same distribution.
At best, this yields a loss of $-2\log(0.5)/n$, which is just below the loss observed in Figure~\ref{fig:axe_loss} when considering the average target sequence length in IWSLT'14 DE-EN is around $\bar{n} \approx 30$.

\paragraph{Inference produces empty sequences}
The model essentially learns to produce the blank token $\varepsilon$ in the first step, and then copy the latest token that is fed into the decoder as input.
During training, that input is indeed the target token.
At inference, however, it is the model's prediction from the previous timestep.
Since the first prediction is $\varepsilon$, the model will continue and predict the blank token until the end of the sequence.

\paragraph{This exploit is not unique to AXE}
AXE is not the only latent alignment objective that the model can ``game'' when coupled with teacher forcing.
We would see a similar phenomenon if we were to use CTC with a longer prediction sequence; for example, if we doubled the prediction length \cite{Libovick2018EndtoEndNN} and applied a version of teacher forcing that feeds each target token twice in a row.
In fact, every latent alignment objective that can align a prediction $p_j$ with a target $y_i$ where $i<j$ will be subject to this exploit, and allow a model trained with teacher forcing to glimpse into the future.

\paragraph{Restricting AXE to causal alignments leads to the trivial alignment}
We further limit AXE to allow only causal alignments, where a prediction $p_j$ may only align with a target $y_i$ if $i \geq j$.
After training with the restricted objective, we observe that AXE selects the trivial alignment ($i=j$) in 98\% of the validation set sentences, whereas the remaining 2\% contain only minor deviations from the trivial alignment, typically one \textit{delimiter} quickly followed by one \textit{clone prediction}.

\section{Conclusion}

This work elaborates why latent alignment objectives are incompatible with autoregressive models trained with teacher forcing.
That said, teacher forcing might not be the best way to train a machine translation model \cite{Bengio2015ScheduledSF, Goyal2016ProfessorFA, Ghazvininejad2020SemiAutoregressiveTI}, and perhaps a future alternative could reopen the discussion on applying latent alignment objectives to autoregressive models.

\section*{Acknowledgements}
This work was supported in part by Len Blavatnik and the Blavatnik Family foundation, the Alon Scholarship, and the Tel Aviv University Data Science Center.

\bibliography{anthology,custom}
\bibliographystyle{acl_natbib}



\end{document}